\documentclass[smallextended]{svjour3}       
\smartqed  
\usepackage{graphicx}
\usepackage{amssymb}
\usepackage{epsfig}
\usepackage{epstopdf}
\usepackage{cite}
\usepackage{caption}
\usepackage[ruled]{algorithm2e}
\begin{document}

\title{FMT:Fusing Multi-task Convolutional Neural Network for Person Search
}


\author{$^1$Sulan Zhai \and $^1$Shunqiang Liu \\
 $^2$Xiao Wang \and $^2$Jin Tang
}


\institute{$^1$School of Mathematical Sciences, Anhui University, Hefei 230601, China\\
$^2$School of Computer Science and Technology, Anhui University, Hefei 230601, China\\
E-mail: sulanzhai@gmail.com
}

\date{Received: date / Accepted: date}

\maketitle

\begin{abstract}
Person search is to detect all persons and identify the query persons from detected persons in the image without proposals and bounding boxes, which is different from person re-identification. In this paper, we propose a fusing multi-task convolutional neural network(FMT-CNN) to tackle the correlation and heterogeneity of detection and re-identification with a single convolutional neural network. We focus on how the interplay of person detection and person re-identification affects the overall performance. We employ person labels in region proposal network to produce features for person re-identification and person detection network, which can improve the accuracy of detection and re-identification simultaneously. We also use a multiple loss to train our re-identification network. Experiment results on CUHK-SYSU Person Search dataset show that the performance of our proposed method is superior to state-of-the-art approaches in both mAP and top-1.
\end{abstract}

\section{Introduction}
\begin{figure*}
  \includegraphics[width=1.01\textwidth]{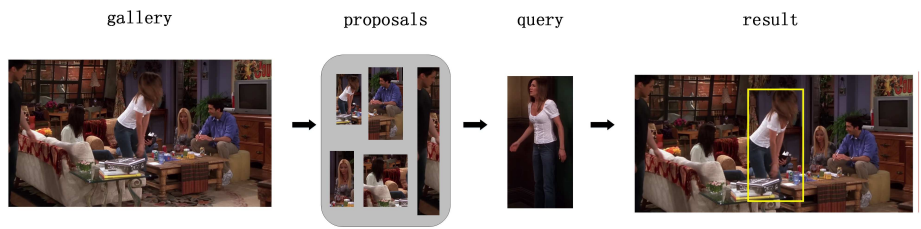}
\caption{Process of person search. For each query person, we detect possible query persons in the gallery images, and then compares all possible pairs of the query to identify the target person.}
\label{fig:1}       
\end{figure*}

Person search combines person detection and person re-identification, which detects all candidate persons in an image and then compares all possible pairs of the query persons to identify the target persons, which is different from person re-identification. It is a challenging and fast-growing field. It has many important applications in video surveillance and multimedia, such as pedestrian retrieval[1] and cross-camera visual tracking[2]. The recent work[3] proposed an end-to-end person search model based on a single convolutional neural network, which adopts proposed Online Instance Matching (OIM) loss function to train re-identification networks, and built a large-scale benchmark dataset for person search (CUHK-SYSU). Person search is generally based on two-stage search strategy. Firstly, we detect all candidate persons in an gallery images. Secondly, we re-identify the query persons from the candiate persons by making a comparison between the query person and all the candidates. See Fig. 1 for demonstration.

In general, a single task needs a single convolutional neural network to extract the specific features. Multi-task learning,where a single network is trained to tackle several related tasks, may learn more robust  presentation and reduce complexity and training time. it has been successfully  used to perform the tasks, such as  person re-identification, face alignment. The recent work[3] introduced the multi-task learning into person search by combining the person detection and person re-identification. However, it only considers re-identification as expansion of detection task. Detection and re-identification  are not only interrelated, but also heterogeneous. They are two different task. So we should consider the heterogeneity of detection and re-identification when designing the multi-task CNN for person search.

In this paper, we propose a fusing multi-task convolutional neural network
(FMT-CNN) to tackle heterogeneity of detection and re-identification and focus on how the interplay of pedestrian detection and person re-identification affects the overall performance.
Person label is one kind of important information that identifies person which will help to improve
the accuracy of person re-identification. In a single convolutional neural network, we add person labels into the RPN and produce more suitable features for re-identification task while keeping the accuracy of detection unchanged. To improve the performance of re-identification network, we use the multiple loss and the more suitable features to train re-identification network. Finally, our proposed FMT-CNN achieves 77.15\% mAP, 79.83\% top-1 accuracy and 90.90\% top-5 accuracy on CUHK-SYSU[3], which shows that FMT-CNN can outperform state-of-the-arts in both
mAP and top-1 evaluation protocols. Our works can be summarized as the following three aspects.(1) We present an efficient fusing multi-task convolution neural network(FMT-CNN) for person search. (2) We add person labels into region proposal network to tackle heterogeneity of detection and re-identification. (3) We adopt the multiple loss to better train re-identification network.
\section{Related Works}

 {\bfseries Person re-identification}[4][5][6][7] aims to match the query person across camera views, which is sub-task of person search. Existing person re-identification (re-id) methods[6][8][9][10][11] target on designing hand-crafted discriminative features, or learning distance metrics to re-identify the target person. Most of works[12][13][14] focus on employing more effective loss function and classifier to improve performance. Deep learning is very extensive. Some works[15][16][17] also propose  person re-identification methods based on deep learning, which design the structure of deep learning network to improve the performance of their methods.
 Multi-task learning is raised because we focus on a single task, and ignore other information that might help optimize metrics. For re-identification sub-task, works[18][19][20][21] always adopt multi-task learning to train their method, which have all achieved good results. Hence we employ multi-task learning to train our network to improve the accuracy.

{\bfseries Person detection}, many works have been made to improve the performance of person detection such as Deformable Part Model (DPM)[22], Aggregated Channel Features (ACF)[23], Checkerboards[24] and etc, which depend on hand-crafted features to detect persons. Recently more works focus on deep learning framework. And then convolution channel features (CCF)[25], Region-CNN (R-CNN)[26] ,Fast R-CNN[27] and Faster R-CNN[28] are proposed , which are deep learning frameworks for object detection and the accuracy of deep learning frameworks are much higher than traditional methods that rely on hand-crafted features. In addition, Faster R-CNN has been employed for person detection.

Recently, [3] proposed an end-to-end deep learning framework for person search to jointly detect and re-identify. Meanwhile, the work built a novel dataset for person search. It used a single convolutional neural network for detecting and re-identifying the target without considering the heterogeneity of two tasks. We employ a fusing multi-task network to jointly detect and identify for person search, which is different from above work[3]. We not only consider the sharing features of the two tasks, but also tackle heterogeneity of detection and re-identification. We add the person labels into RPN to extract the more efficient features.
\section{Proposed Approach}

In this section, we present the details of our proposed fusing multi-task convolutional neural network for person search(FMT-CNN). The overall of our proposed fusing multi-task convolutional neural network is shown in Fig. 2.
\begin{figure*}
  \includegraphics[width=1.05\textwidth]{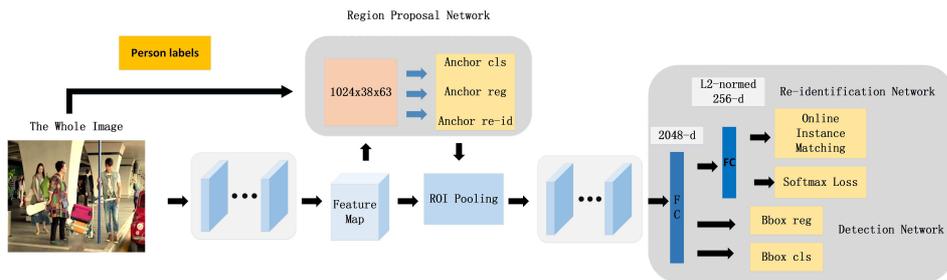}
\caption{The Architecture of FMT-CNN. It consists of detection network, re-identification network and region proposal network. We use ResNet50 for shared feature learning.}
\label{fig:2}       
\end{figure*}
\subsection{Architecture Overview}
Resnet50[29] is used as the backbone network of FMT-CNN, which is a single convolutional neural and consists of three parts: region proposals network(RPN), detection network and re-identification network. Our RPN is different from  stand RPN. We add person labels to extract more useful deep learning features of possible query pedestrians from the whole gallery image in RPN, which can help re-identification network to improve the performance under the same recall and average precision (ap) of the detection network. And we adopt the multiple loss to train re-identification network. The input of our network is a whole image that includes a lot of pedestrians, and then Resnet50[29] extracts the deep learning features of images. The features and the person labels are fed into the RPN together. As a result, the RPN produces more efficient features for identification. The ROI pooling layer encodes all the output of the the RPN into the same size of $14\times14\times1024$ feature cubic. The $14\times14\times1024$ feature cubic turns into the $2048$ dimension feature at the end of Resnet50 through convolution operations. The $2048$ dimension feature is fed into the detection network and re-identification network respectively. Finally, our framework outputs the final search result, which includes the accurate bounding boxes of the query person in the gallery images.
\subsection{Region Proposal Network with Person Labels}
We adopt the Resnet50[29] pre-trained as the backbone network. Our RPN is built on the top of the intermediate convolutional layer of the Resnet50[29]. The standard RPN was proposed in Faster-RCNN, which is a single-category detection network. It produces the possible bounding boxes for detection network. The standard RPN is designed only for detection task. Person search covers detection and re-identification tasks which need different feature. The detection and re-identification tasks need sharing the same feature in a single network, meanwhile they are heterogeneity tasks. We aims to design RPN to fit the two tasks not just to satisfy detection task.

Person labels are the identity information of the labeled pedestrians, which covers the important identification information. Motivated by the finding, we present our RPN with person labels to improve the re-identification accuracy while maintaining the detection accuracy. The input of the standard RPN is the deep features of the image only to predict possible bounding boxes of the target. In our model, RPN with person labels regresses possible bounding boxes from gallery images and respectively predicts that whether possible bounding boxes are the query identities, which produces more useful proposals for re-identification. The distinction between the standard RPN and our RPN is that our RPN's input includes the personal label information.

The RPN can produce better proposals, because person labels are fed into RPN, which includes more information to produce proposals. These proposals are much preciser to the target persons. Hence the PRN with person label produces more useful proposals to handle task heterogeneities. Then we exploit the ROI pooling layer to extract efficient features of these proposals for re-identification and detection. During the training phase, the loss of classification, regression and identification are respectively $\mathcal{L}_{cls}$,$\mathcal{L}_{reg}$ and $\mathcal{L}_{OIM}$. The loss of RPN with person labels is:

\begin{eqnarray}
\mathcal{L}_{rpn}=\mathcal{L}_{cls}+\mathcal{L}_{reg}+\mathcal{L}_{OIM}
\end{eqnarray}

\subsection{Detection and Identification Network}
The RPN with person labels outputs the person proposals from the gallery images, which are fed into the ROI-pooling layer to be normalized into the same size. The ROI-pooling layer outputs features of possible persons. After convolution operations, these features are sent into the detection and re-identification network.

For detection part, we use the smoothed-L1
loss[27] for the bounding box regression layer and a softmax loss to classify whether the possible proposals are the query persons. The output of the Resnet50 is $2048$ dimension deep learning features, which are the input of the detection network. And the output of detection network is the location of the query person in the gallery images. The loss of detection network is defined as:
\begin{eqnarray}
\mathcal{L}_{det}=\mathcal{L}_{cls}+\mathcal{L}_{reg}
\end{eqnarray}
The $\mathcal{L}_{cls}$ is softmax loss, and $\mathcal{L}_{reg}$ is smoothed-L1.

For re-identification part, $2048$ dimension deep learning features are fed into the re-identification network, which are the features of the gallery persons. Firstly we project the features into a L2-normalized 256 dimensional subspace and use them to compute cosine similarities with the query person. Here we employ the multiple loss to train re-identification network, which includes a softmax and an OIM loss to identify whether similar persons of the gallery images are the query person. The OIM loss can maximize the discrepancy among different people to distinguish different people, which is a variation of softmax loss. The softmax loss forces the features of different classes staying apart to better distinguish between identities and predicts false alarms. Hence we design the multiple loss to train network, which is defined as:
\begin{eqnarray}
\mathcal{L}_{re-id}=\mathcal{L}_{1}+\mathcal{L}_{OIM}
\end{eqnarray}
\begin{eqnarray}
\mathcal{L}_{1}=-\log{a_i}
\end{eqnarray}
\begin{eqnarray}
\mathcal{L}_{OIM}=-log{p_i}
\end{eqnarray}

\begin{eqnarray}
a_i=[\frac{exp(w_i^T{x})}{\sum_{j=1}^{K+1}{exp(w_j^T{x})}}]
\end{eqnarray}
\begin{eqnarray}
p_i=[\frac{exp(v_i^T{x/\gamma})}{\sum_{j=1}^K{exp(v_j^T{x/\gamma})}+\sum_{k=1}^Q{exp(u_k^T{x/\gamma})}}]
\end{eqnarray}

Where $x\in{R^D}$ is the input of loss function, and $D$ is dimension of deep learning features, which are the final output of the network. $K$ is the number of the target classes and class $K+1$ is the background. $i$ is class identity. $a_i$ is the probability of sample belonging to class $i$. $W\in{R^{D\times{K}}}$ is weight matrix corresponding to K class. $W^Tx$ is the probability of a sample belonging to each class. Besides $\mathcal{L}_{re-id}$ is the loss of re-identification task. $p_i$ is similar to $a_i$, but it is related not only to the labeled persons, but also to the unlabeled persons, where $\gamma\in{[0,1]}$. $V\in{R^{D\times{K}}}$ stores the features of all the labeled persons. $V\in{R^{D\times{K}}}$ stores the features of all the labeled identities $V^Tx$ is used to compute similarities between the mini-batch sample and all the labeled persons. $U\in{R^{D\times{Q}}}$ is to store the features of unlabeled gallery persons. $Q$ is the size of unlabeled gallery persons. $U^Tx$ is used to compute their similarities with the mini-batch sample.

Our proposed FMT-CNN consists of three parts, which are RPN with person labels, detection network and re-identification network. The $\mathcal{L}_{det}$ is the loss of detection network, and the $\mathcal{L}_{re-id}$ represents the loss of re-identification network. The loss of the RPN is $\mathcal{L}_{rpn}$. Hence the loss function of our proposed FMT-CNN is defined as:
\begin{eqnarray}
\mathcal{L}=\mathcal{L}_{det}+\mathcal{L}_{re-id}+\mathcal{L}_{rpn}
\end{eqnarray}

\subsection{Jointly Training and Testing}
At the training phase, we use softmax loss and OIM loss to jointly train our model. The OIM considers the labels and unlabels. The softmax loss calculates the labels and unlabels. With the increasing of gallery size, unlabeled pedestrians increase and only using OIM will affect the accuracy of the re-identification network. So we adopt softmax loss and OIM loss to jointly train our model. The classifier matrix of softmax suffers from large variance of gradients. We L2-normalize 256 dimensional subspace and propose pre-trained model to reduce the variance of gradients of the classifier matrix to learn softmax. The algorithm of training FMT-CNN is presented in Alg.1.
\begin{algorithm}
            \LinesNumbered
            \caption{Training FMT-CNN}
            \KwIn{a pre-trained ResNet-50, bounding boxes, labels, person labels}
            Add the person labels to the RPN\;
            Fine-tune the network with OIM loss and RPN\;
            Train the fine-tuned network with OIM, softmax loss and RPN\;
        \KwOut{a fine-tuned ResNet-50}
\end{algorithm}

At the test phase, a whole gallery image is fed into the fusing multi-task convolutional neural network for person search. For each gallery image, we use RPN with person labels to get all gallery candidate persons in the gallery image. Meanwhile we use the same network to extract the deep learning features of the query person. Then we extract the deep learning features of all candidate pedestrian to compute the pairwise Euclidean distances between the features of the query person and features of the gallery candidate pedestrians. Finally, we select the shortest distance as the final result.
\section{Experiments}
\subsection{Dataset and Setting}
In order to evaluate our proposed FTM-CNN, we compare FTM-CNN with the proposed baseline method on CUHK-SYSU[3]. CUHK-SYSU is a large scale and scene-diversified dataset for person search, which annotates all the 96,143 pedestrians bounding boxes and 8,432 labeled identities and contains 18,184 images. Because person search requires to identify the target persons mainly according to their clothes and body shapes rather than their faces, people were annotated who do not change clothes and decorations. And every query person appears in at least two images captured. This dataset provides the official training/test split. There are 2,900 query persons and 6,978 images in the test set. There are 11,206 images and 5532 query persons in the training set.

In this paper, we implement our network using Caffe[26] deep learning framework. The convolution neural network we use is Resnet50[29]. For the training of the FMT-CNN, iteration times is 70 thousand for CUHK-SYSU dataset. For each iteration 10000 times, the learning rate decreased to 0.1 per ten thousand times. The initial learning rate we set is 0.001 and decays at the rate of 0.9 for the weight updates. For one gallery image, our proposed framework spends approximately 1s outputting a final searched result on the CUHK-SYSU.

\subsection{Experiment and Analysis}
\begin{table}
\footnotesize
\renewcommand\arraystretch{1.5}
\begin{center}
\caption{\upshape Performance comparison between our FMT-CNN and state of the art methods with 100 gallery size setting. State of the art methods include ACF+person re-id methods, CCF+person re-id methods, CNN+person re-id methods }
\setlength{\tabcolsep}{3mm}{
\begin{tabular}{l|c c}\hline
\textbf{Method} & \textbf{mAP(\%)} & \textbf{top-1(\%)}\\
\hline
ACF[19]+DSIFT+Euclidean &  21.7 & 25.9\\
ACF+DSIFT+KISSME &  32.3 & 38.1\\
ACF+Bow+Cosine &  42.4 &48.4\\
ACF+LOMO+XQDA &  55.5 &63.1\\
ACF+IDNet & 56.5 & 63.0\\
\hline
CCF[21]+DSIFT+Euclidean &  11.3 & 11.7\\
CCF+DSIFT+KISSME &  13.4 & 13.9\\
CCF+Bow+Cosine &  26.9 &29.3\\
CCF+LOMO+XQDA &  41.2 &46.4\\
CCF+IDNet & 50.9 & 57.1\\
\hline
CNN[24]+DSIFT+Euclidean &  34.5 & 39.4\\
CNN+DSIFT+KISSME &  47.8 & 53.6\\
CNN+Bow+Cosine &  56.9 &62.3\\
CNN+LOMO+XQDA &  68.9 &74.1\\
CNN+IDNet & 68.6 & 74.8\\
\hline
JDI-PS[3] & 75.5 & 78.7\\
\textbf{JDI-PS+pl} & \textbf{76.2} & \textbf{79.4}\\
\textbf{FMT-CNN} & \textbf{77.2} & \textbf{79.8}\\
\hline
\end{tabular}}
\end{center}
\end{table}

\begin{figure*}
\centering
\includegraphics[width=1.01\textwidth]{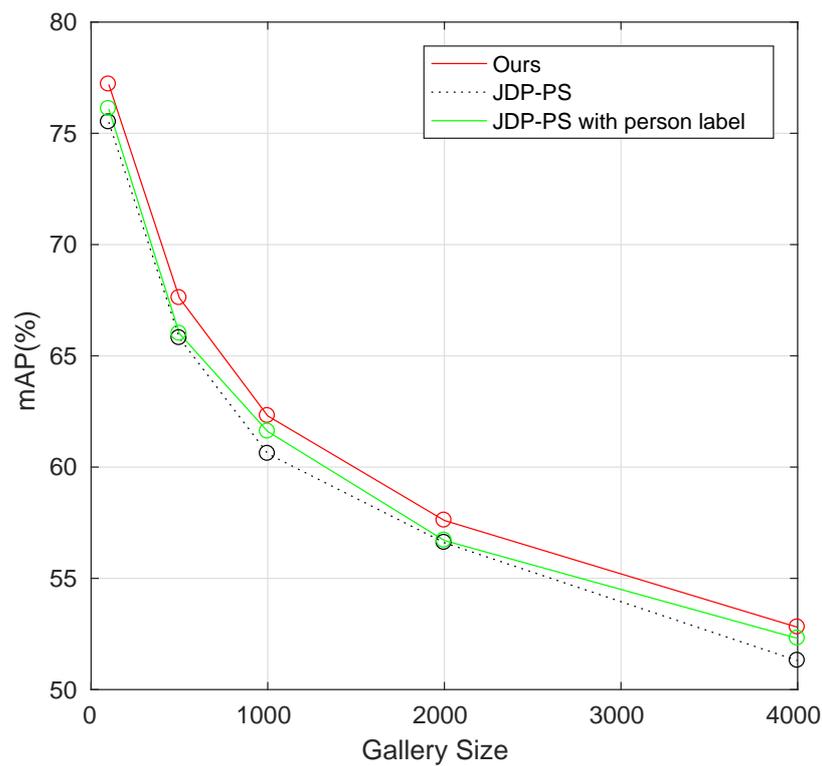}
\includegraphics[width=1.01\textwidth]{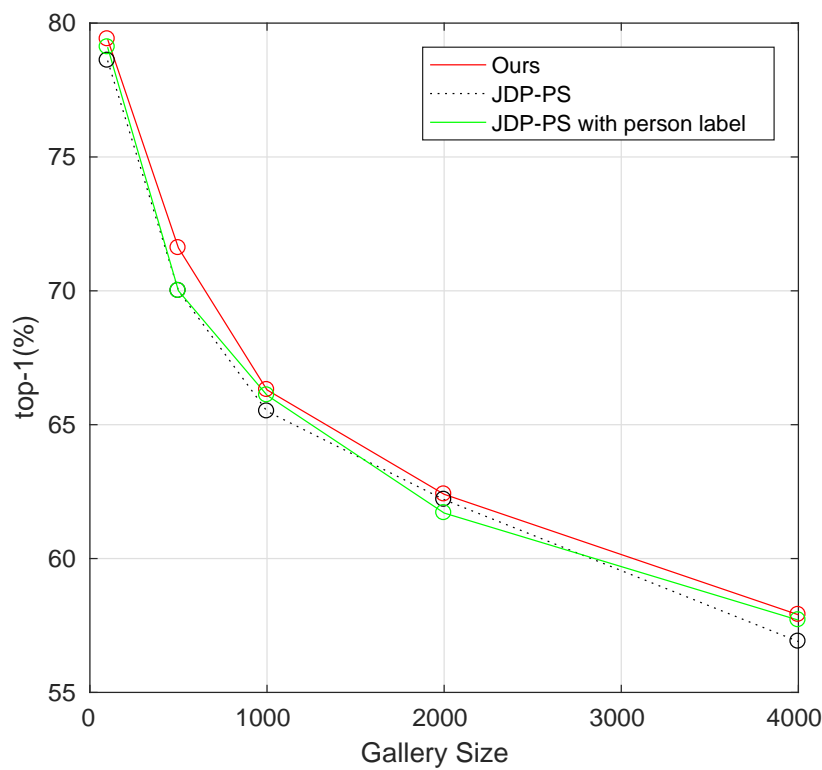}
\caption{Performance curve of JDI-PS, JDI-PS with person label and FMT-CNN. Figure above shows relation between mAP and gallery size. Figure below shows relation between top-1 and gallery size.}
\label{fig:3}
\end{figure*}

\begin{table}
\footnotesize
\renewcommand\arraystretch{1.5}
\begin{center}
\caption{\upshape Performance comparison between ours and JDI-PS on CUHK-SYSU with 100 gallery size setting}
\setlength{\tabcolsep}{4mm}{
\begin{tabular}{l|cccc}\hline

\textbf{search ranking(\%)} & \textbf{mAP} & \textbf{top-1} & \textbf{top-5} & \textbf{top-10}\\
\hline
JDI-PS & 75.42 & 78.48 & 90.07 & 92.34\\
\textbf{Ours} & \textbf{77.15} & \textbf{79.83} & \textbf{90.90} & \textbf{93.28}\\
\hline
\end{tabular}}
\end{center}
\end{table}

We compare our proposed FMT-CNN with several state-of-the-art methods on CUHK-SYSU[3]. And JDI-PS[3] is baseline. For other methods, we compare several popular hand-crafted features(DenseSIFT-ColorHist (DSIFT)[32], Bag of Words (BoW)[33], and Local Maximal Occurrence (LOMO)[34]), two specific distance metric(KISSME[35], and XQDA[34]) and three pedestrian detectors(ACF[23], CCF[25] and their respective R-CNN[28]) with our proposed framework.

We validate the person search performance of our proposed framework on CUHK-SYSU with 100 gallery size setting in Table 1, where CNN denotes Faster-RCNN [28] with Resnet50 and IDNet is the re-identification part in the framework of JDI-PS[3]. It suggests that FMT-CNN boosts all compared methods both the mAPs and top-1 metrics. In order to further verify that the performance of our framework is improved, we compare our FMT-CNN with JDI-PS on mAP, CMC top-1, CMC top-5 and CMC top-10 in Table 2, which provides our framework outperforms JDI-PS. Fig 3 shows that our FMT-CNN surpasses the state-of-the-art JDI-PS[3] in the mAP from different gallery sizes(including 100, 500, 1000, 2000, 4000), under the same recall. As the size of the gallery increases, the mAP and CMC top-1 declines gradually. But in terms of the mAP and CMC top-1, our proposed FMT CNN is nearly 2\% higher than JDI-PS[3].

In addition, Fig 3 shows that the RPN with person labels can achieve a high performance for re-identification task under the same recall. This proves that employing person labels to produce the possible pedestrians of the gallery images can address the heterogeneity of detection and re-identification. And the mAP and top-1 of our proposed FMT-CNN is higher than JDI-PS and JDI-PS with person labels. But as the size of the gallery increases, we find that the performance of FMT-CNN declines gradually. However, the accuracy of our proposed framework, which adopts person labels and multiple loss, is higher than that of JDI-PS in terms of the mAP and top-1 for different gallery sizes.

\section{Conclusion}
In this paper, we propose a novel fusing multi-task convolutional neural network for person search(FMT-CNN) to address the heterogeneity of detection and re-identification. In fact, we employ person labels in the RPN to improve the accuracy of re-identification network by producing the more useful person proposals, while we keep the accuracy of the detection network unchanged. Meanwhile, we adopt the multiple loss to better train re-identification network. Extensive experiments show that our FMT-CNN achieves the state-of-the-art performance.

\section{Acknowledgement}
This work was supported in part by the National Natural Science Foundation of China
under Grant 61872005, in part by the Natural Science Research Project of Anhui universities of China under Grant KJ2018A0029, and in part supported by open project of Anhui University KF2019A03.

\begin{figure}[h]
{\textbf{Sulan Zhai} received the B.S degree and Ph.D.degree in computer  science from Anhui University, Hefei, China, in 2001 and 2008 respectively. She is currently a vice Professor with the school of Mathmatics Science, Anhui University, China.  Her crrent research interests include computer vision, pattern recognition and machine learning.}
\end{figure}

\begin{figure}[h]
{\textbf{Shunqiang Liu} received the B.S. degree in the school of mathematical and statistical at Hefei normal University, Hefei, China, in 2014. He is currently pursuing the M.S. degree in the school of mathematical sciences at Anhui University, Hefei, China. His current research interests mainly about computer vision, machine learning, and pattern recognition.}
\end{figure}

\begin{figure}[h]
{\textbf{Xiao Wang} received the B.S. degree in Western Anhui University, Luan, China, in 2013 and the M.S. degree in Anhui University, Hefei, China in 2016, respectively, where he is currently pursuing the Ph.D. degree in computer science in Anhui University. His current research interests mainly about computer vision, machine learning, and pattern recognition.}
\end{figure}

\begin{figure}[h]
{\textbf{Jin Tang} received the B.Eng. degree in automation and the Ph.D. degree in computer science from Anhui University, Hefei, China, in 1999 and 2007, respectively. He is currently a Professor with the School of
Computer Science and Technology, Anhui University. His current research interests include computer vision,
pattern recognition, and machine learning.}
\end{figure}

\end{document}